\documentclass[runningheads]{llncs}
\usepackage{graphicx}
\usepackage{subcaption}
\begin{document}
\title{Geometry based Machining Feature Retrieval with Inductive Transfer Learning}
\titlerunning{Geometry based Machining Feature Retrieval}

\author{N S Kamal\inst{1},
H B Barathi Ganesh \inst{2},
V V Sajith Variyar\inst{1},\\
V Sowmya\inst{1},
K P Soman\inst{1}
}
\authorrunning{N S Kamal et al.}
\institute{Center for Computational Engineering and Networking (CEN),\\ Amrita School of Engineering, Coimbatore,\\Amrita Vishwa Vidyapeetham, India\\
\email{nskamal007@gmail.com,vv\_sajithvariyar@cb.amrita.edu,\\
v\_sowmya@cb.amrita.edu,kp\_soman@cb.amrita.edu}\\
\and
RBG.AI, Resilience Business Grids LLP, \\ SREC Incubation Center, Coimbatore, Tamil Nadu, India\\
\email{aiss@rbg.ai}\\
}
\maketitle              % typeset the header of the contribution
\begin{abstract}
Manufacturing industries have widely adopted the reuse of machine parts as a method to reduce costs and as a sustainable manufacturing practice. Identification of reusable features from the design of the parts and finding their similar features from the database is an important part of this process. In this project, with the help of fully convolutional geometric features, we are able to extract and learn the high level semantic features from CAD models with inductive transfer learning. The extracted features are then compared with that of other CAD models from the database using Frobenius norm and identical features are retrieved. Later we passed the extracted features to a deep convolutional neural network with a spatial pyramid pooling layer and the performance of the feature retrieval increased significantly. It was evident from the results that the model could effectively capture the geometrical elements from machining features.

\keywords{Machining Feature Recognition\and Inductive Learning\and CAD\and Geometrical Features\and Spatial Pyramid Pooling}
\end{abstract}

\section{Introduction}

Computer-aided process planning (CAPP) helps determine the processing steps required to manufacture a product or its parts. It serves as the connecting link between Computer Aided Design (CAD) and Computer Aided Manufacturing (CAM). Automated machining feature recognition is a critical component in the detection of manufacturing information from CAD models. When it comes to machine parts, features are semantically higher level geometric elements such as a hole, passage, slot etc. Feature recognition is the process of identifying these features from an image or a 3D model of these machine parts.

Approaching Machining Feature Recognition (MFR) as a supervised problem is not feasible for real-time applications in industries. This is because, different industries have their own  data lakes to store the huge pool of 3D CAD models that include different combinations with respect to their requirements. Preparing a supervised corpus and training them to predict large number of classes is a complex and tedious process to achieve. By observing this, we propose an approach that includes inductive transfer learning for geometrical feature extraction[1]. CAD models of machining features[2] were converted to point cloud data and Minkowski engine[19] was used to handle sparse convolutions. Frobenius norm was used for measuring similarity between the extracted geometrical features and later a spatial pyramid pooling layer was added for making the feature matrices to span the same number of spaces. We have detailed the relevant work associated with machining feature retrieval in section 2, the proposed approach is given in section 3 and observed results with the experiments conducted are given in section 4.

\section{Literature Review}

The attempts at integrating  computer-aided process planning (CAPP) with CAD began decades ago[3]. Several approaches to development of feature recognition techniques followed [4,5]. Major approaches being graph-based methods[11,15], hint based methods[12], convex decomposition, volume decomposition[13] and free form features[14]. One common problem addressed by all these approaches was an increase in computational complexity. Introduction of Neural Networks revolutionized feature recognition as it was excellent at recognizing patterns and was able to tolerate noise in input data.

Some of the most influential recent  works on 3D data representation and feature recognition are as 3D ShapeNets[6], VoxNet[7] and PointNet [8]. However, the most recent and important work on machining feature recognition is FeatureNet [2]. The authors have used Deep 3D Convolutional Neural Networks and it could recognize manufacturing features from the low-level geometric data such as voxels with a very high accuracy. To train FeatureNet, a large-scale dataset of 3D CAD models with labeled machining features was constructed by the authors. They achieved a feature recognition accuracy of 96.7 \% for the recognition of single features. 

The recent break-through in geometric feature extraction was using Fully Convolutional Geometric Features by C. Choy et al[1]. The authors present a method to extract geometric features computed in a single pass by a 3D fully-convolutional network. The predictions have a one-to-one correspondence with input image in spatial dimension. The authors have used a UNet structure with skip connections and residual blocks to extract sparse fully-convolutional features. They used the Minkowski Engine to handle sparse convolutions.

Majority of the existing works on structuring 3D models of machining features are based on low-level geometric characteristics, which makes the model localized and  domain specific. Extracting high level features will result in a generalized model that can be applied in various domains. Utilizing fully convolutional neural networks for feature extraction is proven to be significantly faster than conventional methods[1].
 
\section{Materials and Methods}

\begin{figure}[]
\centerline{\includegraphics[scale=0.35]{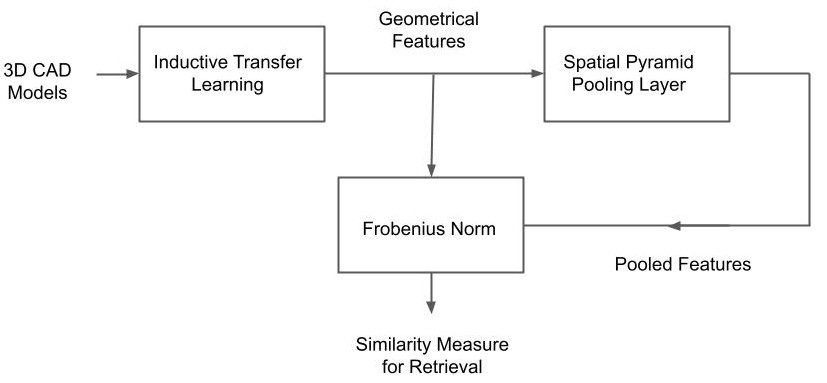}}
\caption{The proposed method for feature extraction and retrieval}
\label{fig:proposal}
\end{figure}

For benchmarking purpose we have taken the synthetic data that includes 24 machining features[2]. Geometrical features were extracted from the CAD models using an inductive transfer learning technique using a model pre-trained with fully convolutional geometric features for the purpose of image registration[1]. Point cloud registration is the underlying source task of this process. The number of extracted geometrical feature vectors from this process was varying with respect to the CAD models. In order to get the same number of feature spaces,later an SPP layer was introduced as the target task of inductive transfer learning process. Frobenius norm was computed to measure the similarity between the CAD models. Based on the obtained similarity score, 3D models were assigned to the respective classes. The overall proposed approach is given in Figure \ref{fig:proposal}.

\subsection{3D CAD Models}
The Dataset used for feature retrieval is adopted from the synthetic dataset generated for FeatureNet [2]. It consists of 24 commonly found machining features in the manufacturing industry (e.g., O-ring, Rectangular passage). Each of the 24 classes has 1000 3D CAD models of varying dimensions and orientations. The features are present in the CAD models as portions removed from a solid cube of dimensions 10cm x 10 cm x10cm (fig:1). The 3D models are available in stereolithography(STL) format, which is a widely used format in the field of computer-aided manufacturing. STL files describe only the surface geometry of a three-dimensional object without any representation of color or  texture. 

\subsection{Methodology}
\subsubsection{Feature extraction with Inductive Transfer Learning}
\begin{figure}
\centering
    \subfloat[\centering ]{{\includegraphics[width=4.15cm]{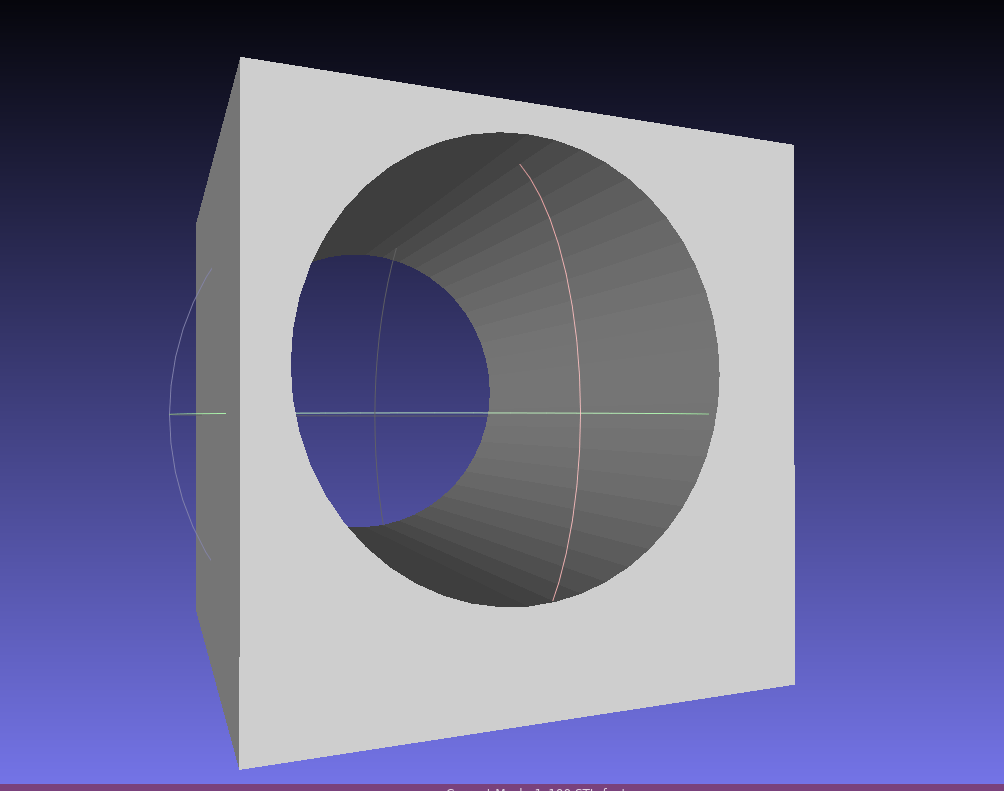} }}%
    \qquad
    \subfloat[\centering ]{{\includegraphics[width=3.9cm]{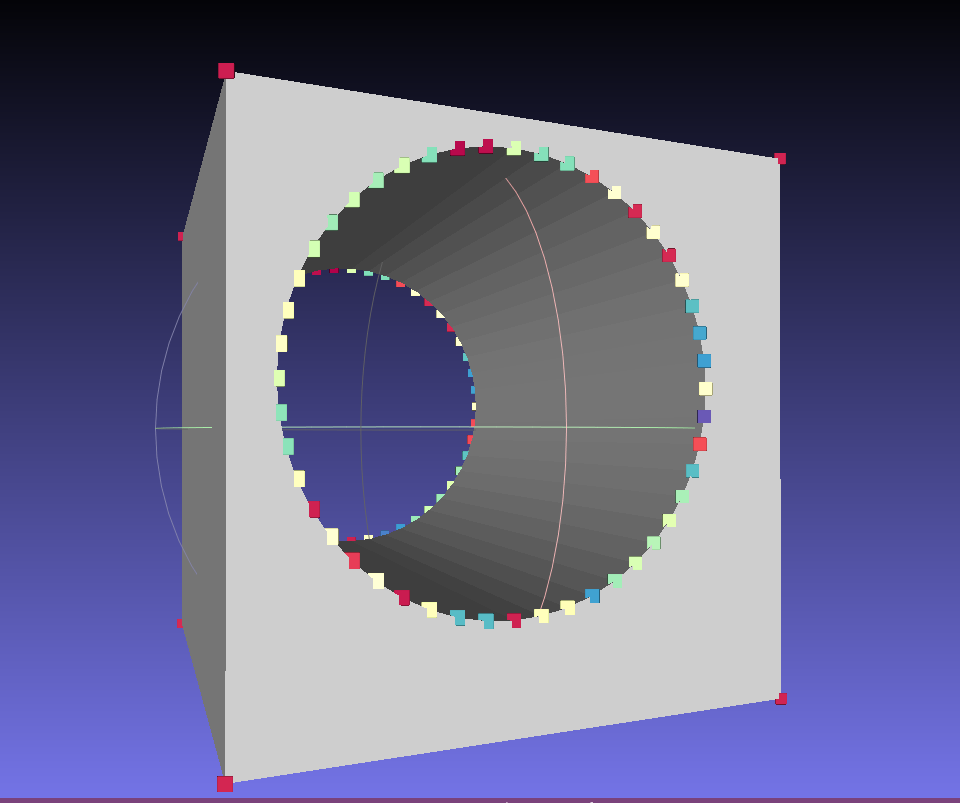} }}%
\caption{(a)A Sample 3D model from class 'through hole', (b) Extracted feature vectors projected to 3D space and laid over the model }
\label{fig:feature_sample}
\end{figure}

The 3D files in STL format were converted into point cloud data with the help of Open3D library[16]. Point cloud data was given as the input to the neural network with ResUNet architecture pre-trained on 3D Match dataset which extracts the high level geometric features [1].The extracted feature vectors were observed to be  32 Dimensional. The number of feature vectors extracted were found to be  different, even for 3D models of the same family. The features were projected to three dimensions with t-Distributed Stochastic Neighbor Embedding(t-SNE)[17] for visualization purposes. A sample 3D model of a through hole with its extracted features laid over it (See Fig: \ref{fig:feature_sample}).

% \begin{figure}
% \centering
% \begin{subfigure}
%   \centering
%   \includegraphics[width= 3 cm]{stl_sample.png}
%   \caption{A subfigure}
%   \label{fig:sub1}
% \end{subfigure}%
% \begin{subfigure}
%   \centering
%   \includegraphics[width= 2.8 cm]{feature_sample.png}
%   \caption{A subfigure}
%   \label{fig:sub2}
% \end{subfigure}
% \caption{A figure with two subfigures}
% \label{fig:test}
% \end{figure}

\subsubsection{Frobenius Norm}
 The extracted feature vectors supposedly should capture the geometrical elements of the machining features. To test this, the extracted features needed to be compared with a similarity measure. For this purpose all the 24000 models were paired with random samples from each class and the Frobenius norm of their feature matrices were found. The Frobenius Norm of a matrix ‘A’ of dimension m x n is defined as the square root of the sum of the squares of the elements of the matrix.

%\begin{align*}
\begin{equation}
\left\|{A}\right\|_F = \sqrt{\sum_{i=1}^{m} \sum_{j=1}^{n} |a_{ij}|^2} 
\end{equation}
%\end{align*}

The difference in frobenius norm obtained for each pair was sorted in descending order along with their corresponding labels. The position of the true label on this list is found out and then the accuracy is calculated. The accuracy and top-five accuracy obtained were 39\% and 86\% respectively. The Top-five accuracy should be satisfactory for retrieval applications, since a set of similar features are returned. From the result  it was evident that the FCGF could capture some geometric information from the 3D models. But the accuracy was not satisfactory enough for real world application.

The reason behind the low performance was later found not to be the extracted geometrical features but the inability of Frobenius norm to serve as a good similarity measure in this particular case. The machining features varied significantly in size and orientation even within the same family. Frobenius norm could only capture the energy of the feature matrix which was not good enough for comparison.   

\subsubsection{Spatial Pyramid Pooling Layer}

\begin{figure}%
    \centering
    \subfloat[\centering ]{{\includegraphics[width=5.2cm]{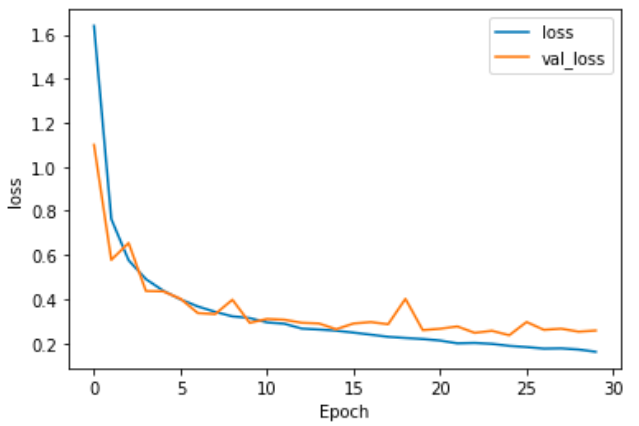} }}%
    \qquad
    \subfloat[\centering ]{{\includegraphics[width=5.2cm]{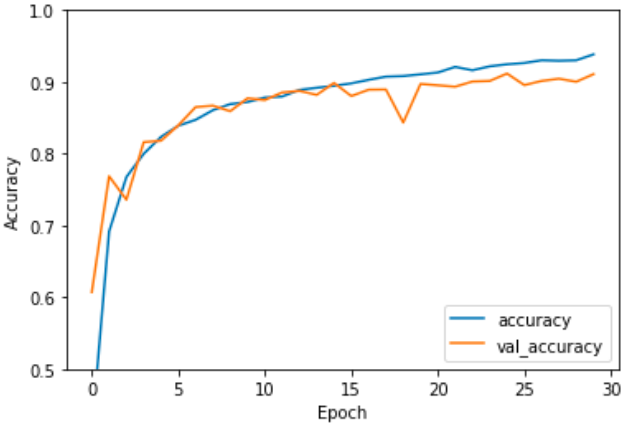} }}%
    \caption{(a).Training and Validation Loss Curves for 30 epochs, (b). Training and Validation Accuracy for 30 epochs}%
    \label{fig:plot}%
\end{figure}

The next target was to improve the results by finding out a better method to capture the similarity between feature matrices of different shapes. A deep neural network with a Spatial pyramid pooling(SPP) layer was used to solve this issue[9,18]. SPP maintains spatial information in local spatial bins of fixed size and removes the fixed-size constraint of the network by giving  an output of a fixed dimension irrespective of input size. The dataset of 24000 models belonging to 24 classes, was split into three subsets – training, validation,testing at  70\%, 15\%, and 15\%, respectively. A CNN architecture  with  97,272 learnable parameters was used to learn the extracted features and in the pre-final layer, Spatial pyramid pooling was introduced as shown in (fig :3). The Frobenius norm is equivalent to the Euclidean norm generalised to matrices instead of vectors. So the Euclidean distance between the output vectors from SPP layer was taken as the measure of similarity between features. Adam optimizer and categorical cross entropy loss was used for training the model for a mere 30 epochs and the loss and accuracy is shown in the graphs (See Fig :\ref{fig:plot}). 

\begin{figure}[htbp]
\centerline{\includegraphics[scale = 0.4]{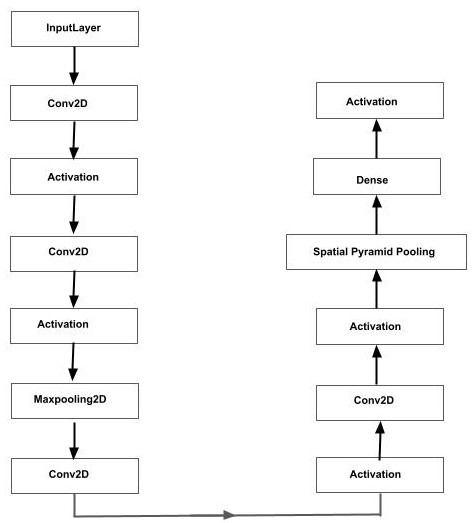}}
\caption{The architecture with SPP layer used for learning extracted  features}
\label{architecture}
\end{figure}

\section{Results and Discussion}

The inclusion of SPP layer imporved the performance of feature retrieval.The testing accuracy obtained  from the current  model  was 86\% and the top-5 accuracy was 95 \% for 30 epochs. Pyramid pooling is robust to object deformations and is suitable for data of higher dimensions. Hence it was able to handle machining features with varying physical dimensions and it proved to be a better method than taking matrix norm of the extracted features directly. As the dimensionality of the data increases, the less useful Euclidean distance becomes. This might be one of the reasons why it didn't work directly on the feature matrices but worked well on the lower dimensional output from the SPP layer.

One major application of this work in real life would be the retrieval of CAD models from the database identical to an input CAD model given by the user. To test the feasibility of this application, some CAD models belonging to families of machining features with different geometrical properties were taken from the test set and we performed a retrieval of the most similar features. The search returned other features from the same family first and then returned features from other geometrically similar families. The categories of the top features returned by sample files from four families of distinct geometrical properties are listed in \ref{tab:Table1}. Family 1 is the most identical one to the family of test feature and family 2 is the second most identical one.

\begin{figure}%
    \centering
    \subfloat[\centering a]{{\includegraphics[width=3cm]{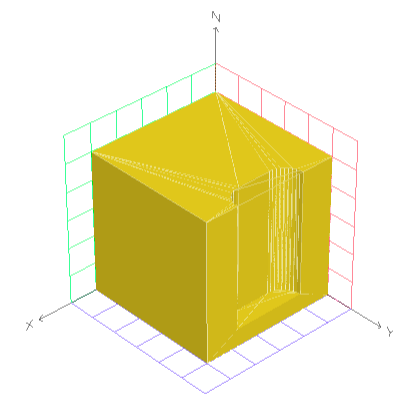} }}%
    \qquad
    \subfloat[\centering b]{{\includegraphics[width=3.2cm]{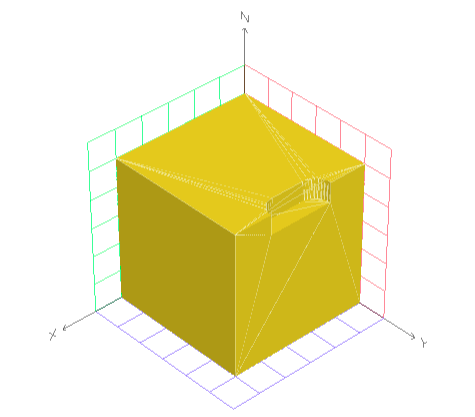} }}%
    \caption{[a].Test model - ID: 1990, Family: Circular end blind slot, [b]. Result no.1 - ID: 1214, Family: Circular end blind slot}%
    \label{fig:cad_models}%
\end{figure}

The models retrieved were rendered and inspected for geometric similarities. Searching with circle based features returned CAD models from other circle based families(e.g., O-ring returned circular end pocket and blind hole). The same was the case with rectangular,triangular and 6 sided features. From the output, it was clear that the extracted feature vectors could successfully capture the geometric properties of the machining features. It became evident that the similarity in size or orientation was preceded by the similarity in geometrical features. Each CAD model in the dataset was given a numerical identification number for convenience. The indices of the top retrieved features along with the corresponding Euclidean distances for a test file from the family 'circular end blind slot' is shown in Table \ref{tab:Table2}. We can observe that all the top 5 features belong to the same family as the test file in this case. The test file and the top retrieved CAD model are rendered in Figure \ref{fig:cad_models}. We can observe that the two features are of different size but the network could still capture the similarity between them.

\begin{table}[htbp]
{
\caption{Sample files from four families chosen for testing and the top 2 families retrieved from database }
\begin{center}
\begin{tabular}{|c|c|c|c|}
\hline
\textbf{Family of Test file} & \textbf{Family 1} & \textbf{Family 2}\\
\hline
O-ring & Circular end pocket & Blind hole \\
\hline
Rectangular passage & Rectangular pocket & Rectangular blind step\\
\hline
Triangular pocket  & Triangular passage & rectangular blind step \\
\hline
6 Sides passage &  6 Sides pocket & Circular blind step  \\
\hline
\end{tabular}
\end{center}
\centering
\label{tab:Table1}
}
\end{table}

\begin{table}[htbp]
\caption{Top-5 models retrieved for a sample 3D model of the family circular end blind spot (ID: 1990)}
\begin{center}
\begin{tabular}{|c|c|c|c|c|c|}
\hline
%\multicolumn{4}{|c|}{First-5 similar models of the test file } \\
\textbf{No.} & \textbf{Model ID} & \textbf{Family}  & \textbf{Euclidean Distance}  \\
\hline
1 & 1214  & Circular end blind spot & 8.17 \\
\hline
2 & 1561  & Circular end blind spot & 8.28 \\
\hline
3 & 1326  & Circular end blind spot & 8.49 \\
\hline
4 & 1853  & Circular end blind spot & 8.89 \\
\hline
5 & 1726  & Circular end blind spot & 8.94 \\
\hline
\end{tabular}
\end{center}
\label{tab:Table2}
\end{table}

% Test model - ID: 1990, Family: Circular end blind slot

\section{Conclusion and Future work}
The results indicate that the inductive transfer learning model performed with fully convolutional neural networks, could capture and learn the geometric similarity between 3D models of machining features. Even though the network was pre-trained on 3D image data, it performed well on the 3D CAD model data.  Spatial pyramid pooling proved to be very efficient in handling feature vectors of varying sizes. The model was good enough to perform feature retrieval with 95 \% top-5 accuracy and the retrieval of CAD files  from databases proved to be successful enough for practical applications. This work could be extended for recognition and retrieval of multiple features present in a single CAD model. This work could also be extended to a generalised shape search application.

\end{document}